\documentclass{article}

\PassOptionsToPackage{numbers,sort&compress}{natbib}

\usepackage[preprint]{neurips_2026}

\usepackage[utf8]{inputenc}
\usepackage[T1]{fontenc}

\usepackage[hidelinks]{hyperref}
\usepackage{url}
\usepackage{booktabs}
\usepackage{amsmath}
\usepackage{amssymb}
\usepackage{microtype}
\usepackage{xcolor}
\usepackage{xspace}
\usepackage{graphicx}
\usepackage{enumitem}
\usepackage{needspace}

\newcommand{\method}{\textsc{TwinCheck}\xspace}
\newcommand{\gpt}{GPT-5.6 Sol\xspace}
\newcommand{\clip}{\operatorname{clip}}
\newcommand{\best}[1]{\textbf{#1}}

\title{TwinCheck: Evidence-Grounded Negative-Twin Verification for Stateful Tool Agents}

\author{
  Jiaxuan Dai\\
  Phillips Exeter Academy\\
  \texttt{jdai2@exeter.edu} \\
  \And
  Tianyi Huang\\
  Ryquo\\
  \texttt{tianyi@ryquo.com} \\
}

\begin{document}

\maketitle

\begin{abstract}
A single locally plausible tool call can derail an otherwise successful agent trajectory. Suspicion alone does not justify intervention, because the replacement itself can introduce the very failure verification is meant to prevent. We introduce \method, an inference-time verification policy that considers replacement only when the trace satisfies an evidence condition tied to a local failure hypothesis. It constructs a trace-grounded counterfactual alternative, a \emph{negative twin}, and replaces the actor's proposal only if the twin passes structural checks and the pairwise verifier prefers it in both candidate orders. For paired evaluation, exact replay holds the actor's parsed responses and actions fixed until the first accepted replacement, separating intervention effects from resampling. On 159 primary BFCL V4 multi-turn tasks, the complete policy raises \gpt task success from 45.3\% to 58.5\%, a gain of 13.2 percentage points (95\% task-bootstrap CI [8.2, 18.8]), with 21 rescues and no observed success-to-failure regressions. Across the completed primary set and a conditional extension, all 37 rescues follow the same inverse-action repair pattern. The verifier accepts every constructed candidate and adds no filtering on these trajectories. \method frames execution-boundary repair as a constrained comparison, making the proposed replacement itself the object of verification.
\end{abstract}

\section{Introduction}

Language-model agents increasingly act through external tools rather than produce a single answer. A stateful agent must connect each proposed call to the current request, the environment state, and the results of earlier calls. ReAct and related systems make these intermediate decisions explicit \citep{yao2023react}, while BFCL shows that multi-turn function calling remains difficult even when single-turn calls are strong \citep{patil2025bfcl}. A trajectory can therefore fail because of one small decision, such as undoing a successful prerequisite that the user never asked to reverse.

We call the tool-using agent whose proposed action is subject to verification the \emph{actor}. A natural response is to place an LLM verifier around the actor. Existing work revises model outputs through self-feedback, external evidence, or learned verification \citep{madaan2023selfrefine,shinn2023reflexion,gou2024critic,cobbe2021training,lightman2024verify,zhang2025generative}. The verifier, however, can introduce its own error. Self-correction without new evidence can degrade correct reasoning \citep{huang2024cannot}, and model judges exhibit systematic biases, including sensitivity to candidate order \citep{zheng2023judge,shi2025position}. This risk is asymmetric at the execution boundary: declining a repair preserves the actor's proposal, whereas an unnecessary repair can turn success into failure. Intervention should therefore require a specific, trace-supported account of what the replacement would fix.

\method addresses this precision requirement with a negative-twin protocol. A \emph{negative twin} is a trace-grounded alternative that addresses a local failure hypothesis: it is a candidate repair rather than an intentionally corrupted action. Three trace-local evidence conditions constrain candidate construction, either determining an edit or restricting a model-generated alternative. After structural validation, a verifier compares the actor's proposal and its twin in both candidate orders within one request. Both orientations must support replacement under a gate frozen before testing. Actor-proposed final responses and clarifications are not revised.

These operational evidence conditions do not prove that an action is wrong or recover the user's intent. We evaluate the resulting policy through exact replay, fixing the actor's parsed responses and actions until the first accepted replacement. On 159 primary BFCL V4 tasks, the complete policy improves task success by 13.2 percentage points. The component audit identifies a specific source of this gain: across the completed primary set and conditional extension, all 37 rescues follow removal of the same vehicle-control inverse action, and the verifier accepts all 71 constructed candidates.

We make three contributions:
\begin{enumerate}[leftmargin=1.4em,itemsep=1pt,topsep=2pt]
  \item A selective tool-action repair policy whose evidence conditions constrain the candidate and whose counterbalanced verifier governs replacement.
  \item An exact-replay evaluation that fixes evaluator-visible actor behavior until the first replacement and measures both rescue and harm.
  \item A component audit that separates candidate construction from verifier selection and identifies the repair pattern responsible for the observed gains.
\end{enumerate}

\section{Related work}

\paragraph{Verification and model judging.}
Learned verifiers rank candidate solutions, while process supervision moves feedback from final outcomes to intermediate reasoning steps \citep{cobbe2021training,lightman2024verify}. Generative verifiers instead cast reward modeling as next-token prediction, allowing verification to use generated rationales and additional inference-time computation \citep{zhang2025generative}. For stateful tool-using agents, however, a model judge is an imperfect source of evidence: its decisions can reflect systematic judging biases \citep{zheng2023judge}, including candidate-order effects \citep{shi2025position}. \method therefore consults a pairwise judge only after the trace supports a specific error hypothesis and a structurally valid alternative exists, and it requires the preference to persist when candidate order is reversed.

\paragraph{Feedback and agent repair.}
Feedback-based repair ranges from repeated revision in Self-Refine to cross-trial verbal memory in Reflexion and tool-grounded critique in CRITIC \citep{madaan2023selfrefine,shinn2023reflexion,gou2024critic}. Analyses of intrinsic self-correction nevertheless caution that revision without an external signal can leave reasoning unchanged or make it worse \citep{huang2024cannot}. At the agent-system level, MAST studies recurrent failures in multi-agent traces \citep{cemri2025mast}, while DoVer tests suspected causes through targeted re-execution \citep{ma2026dover}. CausalFlow likewise intervenes on failed traces to attribute failure and produce validated counterfactual repairs \citep{bonagiri2026causalflow}. \method addresses a more specific target: before the next tool action executes, it asks whether observable trace evidence justifies one local substitution.

\paragraph{Selective safeguards for tool agents.}
GuardAgent checks agent actions against executable safety policies, whereas AttriGuard uses counterfactual replay to test whether a tool call depends on untrusted observations \citep{xiang2025guardagent,he2026attriguard}. SABER concentrates oversight on state-changing actions before execution \citep{cuadron2025saber}, and Verify What Matters prioritizes verification when estimated downstream harm is high \citep{tang2026verify}. \method instead starts from a trace-local failure hypothesis and asks whether one evidence-conditioned alternative should replace the proposal.

\paragraph{Tool-agent evaluation.}
BFCL covers function calling from single-turn invocation through stateful multi-turn tasks; for the latter, its official evaluator applies state- and response-based checks after each turn \citep{patil2025bfcl}. Our experiments use four stateful BFCL V4 categories and the official evaluator.

\section{Problem formulation}

At step $t$ of task $j$, the visible context $c_{j,t}$ contains the user request, trace prefix, tool schemas, and observed tool outputs. An actor proposes action $a_{j,t}$, which may be one or more tool calls, a clarification, or a final response. A wrapper either executes $a_{j,t}$ or substitutes a counterfactual action $\tilde a_{j,t}$. The environment evaluator returns final success $Y_j\in\{0,1\}$.

A switch score $g_{j,t}\in[0,1]$ is nonzero only when the proposal satisfies an evidence condition, a structurally valid twin exists, and the pairwise verifier reaches consensus. Intervention occurs when $g_{j,t}\geq\gamma$, where $\gamma$ is frozen before test evaluation. Intervention quality is measured by the task-level paired outcomes
\begin{align}
\text{rescue} &:\quad Y_j^{\text{actor}}=0,~Y_j^{\method}=1,\\
\text{harm} &:\quad Y_j^{\text{actor}}=1,~Y_j^{\method}=0.
\end{align}
The objective is to produce more rescues than harms while changing as few actions as necessary. A separate trace-anomaly score $d_{j,t}\in[0,1]$ supports descriptive failure-ranking diagnostics; it has no standalone intervention threshold.

\Needspace{28\baselineskip}
\section{Method}
\label{sec:method}

\subsection{Trace evidence and repair eligibility}

The \emph{current-turn trace} contains parsed tool calls and available outputs after the latest user message. The new actor proposal is assessed separately, before execution. The ledger retains every parsed record, including calls with missing outputs, together with its position and inferred outcome.

Repair eligibility is limited to three evidence conditions. We use \emph{certificate} as shorthand for such an operational eligibility condition (Appendix~\ref{app:certificate-examples}).

\paragraph{Lexical inverse-action condition.}
The proposed and earlier successful tool names have the same tokenized object suffix and opposite verbs in a fixed table, while the latest user message contains none of a fixed set of request phrases for the proposed inverse. The predicate also recognizes one or more Boolean flips within the same state-changing tool when all other shared arguments are unchanged.

\paragraph{Repeat-after-error condition.}
The proposal repeats the same tool and arguments after a failed result. Any later successful current-turn tool record suppresses this condition, regardless of tool identity.

\paragraph{Unique-stale-argument condition.}
Exactly one scalar under a target-related argument key matches a value in the history before the latest user message, while that message contains exactly one distinct value of the same recognized type that differs from the earlier value.

The three conditions have severities $1.00$, $1.00$, and $0.95$, respectively. All exceed the implementation's severity floor of $0.90$, which therefore adds no filtering among these types. Actor-proposed final responses and clarifications may be scored but are never revised.

\paragraph{Trace-anomaly score.}
A fixed heuristic maps the trace history and current proposal to $d_{j,t}$. The score increases with repeated calls, observed tool errors, and schema violations, and incorporates the strongest matched rule. It is supplied as context to the verifier, but is not interpreted as a failure probability or directly thresholded to authorize repair. The score, evidence predicates, and policy constants were frozen before either test set was sampled. Section~\ref{sec:ranking} evaluates failure ranking separately from intervention selectivity.

\subsection{Constructing and validating the twin}

The certificate defines the local failure hypothesis and specifies a deterministic edit or constrains a model-generated alternative. Under the unique-stale-argument condition, the sole current value replaces the stale value and every other argument remains unchanged. A lexical inverse-action match deletes every matched call. If no calls remain and the ledger contains at least one success with no unresolved failed current-turn call, \method emits the fixed response ``The requested operation is complete.'' If deletion would leave no calls but the completion condition fails, the unique-stale-argument rule may still supply an edit under its own eligibility checks.

If no deterministic candidate is available, a separate low-effort call to the same underlying model generates an alternative. This generator also handles repeat-after-error cases. It receives the full available message trace, up to eight tool schemas retained for that decision, and an instruction containing the certificate reason and evidence. It does not receive the serialized actor action, although that evidence may disclose the implicated tool or value.

Every nonidentical twin must pass structural checks before judgment. Tool-call twins must satisfy the active JSON schema. Action kind and call count are preserved unless the evidence condition authorizes structural change, as inverse-action deletion and repeat-after-error repair may require. Without that authorization, a model-generated twin must remain within a distance of $0.35$ from the actor action, measured as one minus the Python \texttt{SequenceMatcher} ratio over canonical JSON serializations. For a model-generated tool twin of the same kind, each newly introduced non-Boolean scalar, including a changed tool name, must also occur in the serialized evidence beginning with the latest user message. A failed check leaves the actor action unchanged.

\begin{figure}[t]
  \centering
  \includegraphics[width=\linewidth]{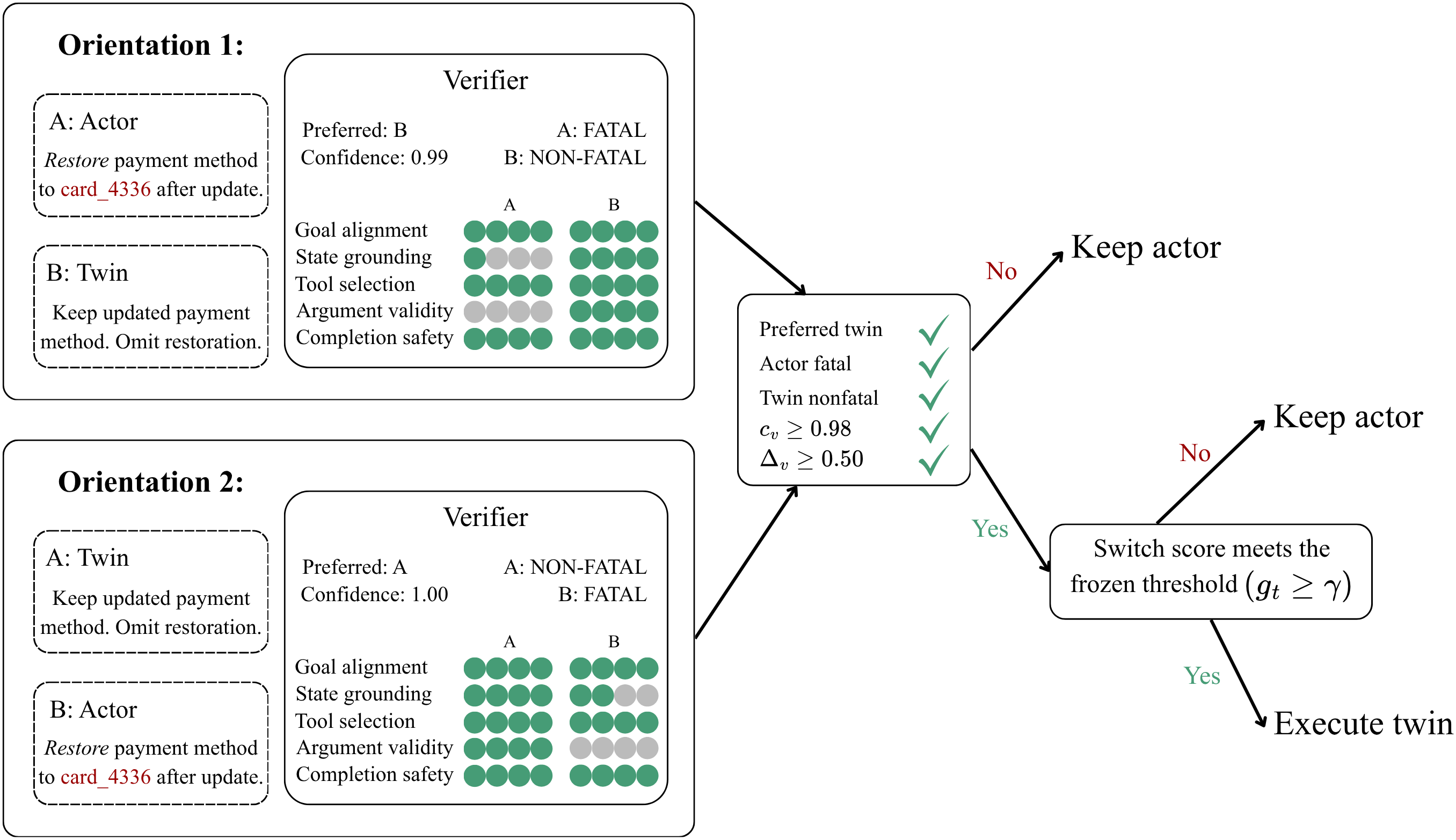}
  \caption{Counterbalanced verification and conservative switching. After evidence and structural checks, the same actor--twin pair is evaluated in both A/B orders within one request. Green circles show the 0--4 rubric scores. Both orientations must satisfy Equation~\ref{eq:switch}, and the switch score must meet the frozen threshold. Otherwise the actor is retained. Actions and scores are schematic; actor/twin tags are explanatory and are not shown to the judge. Task indices are omitted.}
  \label{fig:counterbalanced-verification}
\end{figure}

\subsection{Counterbalanced contrastive verification}

For an eligible pair $(a_{j,t},\tilde a_{j,t})$, the verifier receives a compact recent trace, up to eight tool schemas, the certificate, and its supporting evidence (Figure~\ref{fig:counterbalanced-verification}). The A/B labels contain no explicit actor/twin tag. One structured-output request contains both $(A=a_{j,t},B=\tilde a_{j,t})$ and the reversed order. In each orientation $v$, it reports a preference and a confidence score $c_v\in[0,1]$, assigns each candidate a fatal-error label, and returns 0--4 scores for goal alignment, state grounding, tool selection, argument validity, and completion safety. These are model-reported annotations, not calibrated probabilities. Let $s_{v,k}(u)$ denote action $u$'s score on dimension $k$ and $f_v(u)\in\{0,1\}$ its fatal-error indicator. The normalized quality is
\begin{equation}
q_v(u)=\clip_{[0,1]}\!\left(\sum_k w_k\frac{s_{v,k}(u)}{4}-0.45f_v(u)\right),
\label{eq:quality}
\end{equation}
with weights $(0.25,0.25,0.20,0.20,0.10)$ in the order listed above. Define $\Delta_v=q_v(\tilde a_{j,t})-q_v(a_{j,t})$. If $E_{j,t}\in\{0,1\}$ denotes satisfaction of the evidence and structural gates, the switch score is
\begin{equation}
 g_{j,t}=
 \begin{cases}
 \displaystyle\min_v\min\{c_v,\Delta_v\}, & \text{if $E_{j,t}=1$ and every orientation prefers $\tilde a_{j,t}$,}\\[-1mm]
 & \quad\text{marks $a_{j,t}$ fatal and $\tilde a_{j,t}$ nonfatal, with}\\[-1mm]
 & \quad c_v\geq0.98\text{ and }\Delta_v\geq0.50;\\
 0, & \text{otherwise.}
 \end{cases}
\label{eq:switch}
\end{equation}
Here $v$ ranges over the two candidate orders. The gate requires agreement between the two reported orientations. Both are produced in one request to the same model and are not independent votes. A parseable response that omits or duplicates an orientation yields $g_{j,t}=0$. A schema-invalid or empty structured response triggers one additional verifier query; if that parse also fails, the outer wrapper retains the actor action and logs an operational exception.

\subsection{Calibration, intervention, and exact replay}

Calibration uses a task-ID-disjoint shadow run with no action replacement. For each calibration task, $G_j=\max_t g_{j,t}$, and the actor outcome $Y_j^{\mathrm{actor}}$ is recorded. When at least one actor trajectory succeeds, the threshold is
\begin{equation}
\gamma=\max\!\left\{0.50,~\operatorname{nextafter}\!\left(\max_{j:Y_j^{\mathrm{actor}}=1}G_j,+\infty\right)\right\},
\label{eq:calibration}
\end{equation}
where $\operatorname{nextafter}(x,+\infty)$ is the smallest representable value above $x$. Calibration aborts if no actor trajectory succeeds or the resulting threshold exceeds $1$. In the reported 32-task calibration, 17 trajectories succeed and their maximum $G_j$ is $0$, so the prespecified floor remains $\gamma=0.50$. Every nonzero score in Equation~\ref{eq:switch} is already at least $0.50$. The shadow run therefore checks the frozen operating point but adds no filtering beyond that rule.

At test time, \method accepts at most one switch per trajectory. Missing or incompatible calibration stops the repair-enabled run before verifier or twin-generation inference; other operational exceptions retain the actor action and are logged. For evaluation, the actor trajectory is generated once. The wrapper replays the same evaluator-visible parsed responses and actions until a switch is accepted, after which downstream behavior may diverge. A trajectory without an accepted switch must preserve that entire sequence exactly.

\section{Experimental setup}
\label{sec:setup}

\paragraph{Benchmark and splits.}
We evaluate the four scored stateful multi-turn categories in BFCL V4: \nolinkurl{multi_turn_base}, \nolinkurl{multi_turn_miss_func}, \nolinkurl{multi_turn_miss_param}, and \nolinkurl{multi_turn_long_context}. We pin \texttt{bfcl-eval==2026.3.23}, which contains 200 tasks per category \citep{patil2025bfcl}. The upstream scoring configuration excludes \nolinkurl{multi_turn_composite}.\footnote{\href{https://github.com/ShishirPatil/gorilla/blob/6ea57973c7a6097fd7c5915698c54c17c5b1b6c8/berkeley-function-call-leaderboard/bfcl_eval/constants/category_mapping.py}{BFCL category configuration, pinned source snapshot.}} After each turn, BFCL applies state- and response-based checks, and its evaluator never receives judgments from \method. We exclude all 401 task IDs used during engineering or method development. Category-stratified sampling from the remaining IDs produces 32 calibration tasks (8 per category) and a planned 160-task primary set (40 per category).

One prespecified primary task lacked an actor trajectory after a transient API error exhausted the original retry cap, so its exact-replay pair was unavailable. The primary analysis uses the 159 complete pairs. A later run regenerated the missing actor trajectory and replayed it exactly through \method, giving the 160-task \emph{completed-primary sensitivity}.

After completing the missing primary pair, a script launched the extension only if the completed primary set had a positive success gain, zero observed harms, valid exact replay with no wrapper exception, detector ROC-AUC at least $0.65$, and projected cumulative API cost below \$195. All checks passed, and the script sampled 160 tasks, excluding every development, shadow, and primary task ID. The scoring rule, threshold, and repair policy remained fixed. This \emph{conditional extension} is not an independently initiated confirmatory replication: its results and the pooled intervals are descriptive, conditional on the outcome-dependent launch.

\paragraph{Models and execution.}
The actor, pairwise verifier, and model-based twin generator use \gpt via the \texttt{gpt-5.6} API alias and OpenAI Python SDK 2.53.0; the original evaluation was conducted in August 2026. Actor reasoning effort is high, verifier effort medium, and twin-generation effort low. Verifier judgments use structured outputs over the compact context described in Section~\ref{sec:method}. At most two eligible decision points on a trajectory may reach the verifier, and at most one action may be replaced. The actor scores apply to this configuration and split, not the overall BFCL leaderboard.

\paragraph{Baselines.}
We compare with the unwrapped actor and two trigger-matched controls on the completed-primary sensitivity. Neither control receives TwinCheck's candidate. \emph{Selective self-revision} uses the same evidence conditions and replay protocol, requests one direct model-generated action revision, and accepts a changed, schema-valid action that passes the structural checks. The \emph{selective absolute verifier} applies the weighted rubric in Equation~\ref{eq:quality}. For single-candidate rubric quality $q$ and confidence $c$, its health is $h=(2q-1)c$, or $-1$ after a schema violation. It triggers when $h<0.60$, generates a repair from the diagnosis, and accepts it only if it passes the structural checks, no longer triggers, and raises health by at least $0.02$. Its $0.60$ threshold is fixed independently of TwinCheck's shadow calibration. These controls compare repair policies; they are not matched-candidate verifier ablations or implementations of the cited repair systems.

\paragraph{Metrics and uncertainty.}
The primary outcome is the unweighted mean of category-level task success. Percentage-point differences are computed from paired binary outcomes before display values are rounded. We obtain 95\% CIs from 20,000 category-stratified percentile bootstrap resamples, drawing task pairs independently within category, and report exact two-sided McNemar tests \citep{efron1993bootstrap,mcnemar1947note}. These intervals capture variation across the observed tasks conditional on one stored actor trajectory per task. For a dependence sensitivity, we group task IDs by the final decimal integer after the last underscore. These \emph{numeric-suffix clusters} provide only a reproducible proxy for scenario similarity. We resample them jointly across categories and recompute the equal-weight category mean (20,000 resamples; seed 20260818). Detector evaluation uses ROC-AUC and average precision for ranking actor failures, together with precision, recall, and the false-positive rate for selecting them. Its pooled AUC interval uses 20,000 resamples of the 173 numeric-suffix clusters represented among the 320 tasks.

\section{Results}
\label{sec:results}

\paragraph{Task success.}
On the 159 available primary pairs, success increases from 45.3\% for the \gpt actor to 58.5\% with \method, a gain of 13.2 percentage points (95\% task-bootstrap CI [8.2, 18.8]; McNemar $p=9.54\times10^{-7}$). There are 21 rescues and no observed harms. Resampling the 122 numeric-suffix clusters instead of individual tasks gives a 95\% sensitivity interval of [7.8, 19.2] percentage points. Completing the missing pair leaves both conditions unsuccessful, yielding 45.0\% versus 58.1\% in the completed-primary sensitivity; its cluster sensitivity interval is [7.8, 19.1].

\begin{table}[htbp]
\centering
\caption{Exact-replay task success, averaged equally across categories. Differences are percentage points; intervals are 95\% CIs from category-stratified paired bootstrap resampling of tasks.}
\label{tab:main-results}
\small
\setlength{\tabcolsep}{3pt}
\begin{tabular}{lrrrrrr}
\toprule
Test set & $n$ & Actor (\%) & \method (\%) & $\Delta$ & 95\% CI & Rescue / harm \\
\midrule
Primary analysis & 159 & 45.3 & 58.5 & +13.2 & [8.2, 18.8] & 21 / 0 \\
Completed-primary sensitivity & 160 & 45.0 & 58.1 & +13.1 & [8.1, 18.8] & 21 / 0 \\
Conditional extension & 160 & 48.1 & 58.1 & +10.0 & [5.6, 15.0] & 16 / 0 \\
\midrule
Pooled & 320 & 46.6 & 58.1 & +11.6 & [8.1, 15.3] & 37 / 0 \\
\bottomrule
\end{tabular}
\end{table}

\begin{figure}[t]
  \centering
  \includegraphics[width=\linewidth]{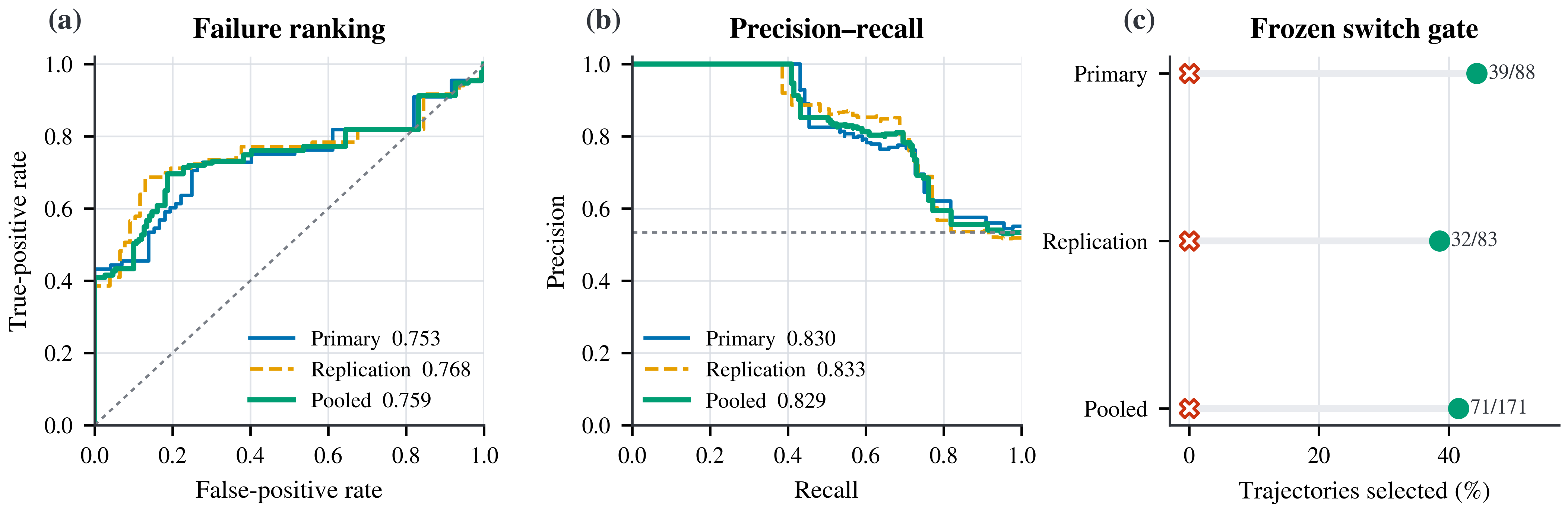}
  \caption{Failure ranking and frozen-gate selectivity on the completed-primary sensitivity (labeled ``Primary''), the conditional extension (labeled ``Replication''), and their pooled set. Extension and pooled results are descriptive, conditional on the outcome-dependent launch described in Section~5. (a) ROC and (b) precision--recall curves. Pooled ROC-AUC is 0.759 (numeric-suffix-cluster 95\% CI [0.686, 0.824]) and average precision is 0.829; the dotted line marks the 0.534 pooled failure prevalence. (c) The switch gate selects actor failures (green) and no actor successes (red) in these sets.}
  \label{fig:detector}
\end{figure}

\begin{table}[htbp]
\centering
\caption{Pooled results by BFCL category (80 tasks each). Intervals use the same paired task-bootstrap procedure and condition on extension launch.}
\label{tab:category}
\small
\begin{tabular}{lrrrr}
\toprule
Category & Actor (\%) & \method (\%) & $\Delta$ & 95\% CI \\
\midrule
Multi-turn base & 55.0 & 65.0 & +10.0 & [3.8, 16.2] \\
Missing function & 52.5 & 65.0 & +12.5 & [6.2, 20.0] \\
Missing parameter & 36.2 & 45.0 & +8.8 & [3.8, 15.0] \\
Long context & 42.5 & 57.5 & +15.0 & [7.5, 23.8] \\
\bottomrule
\end{tabular}
\end{table}

On the conditional extension, success increases from 77/160 (48.1\%) to 93/160 (58.1\%), a gain of 10.0 percentage points (95\% task-bootstrap CI [5.6, 15.0]; $p=3.05\times10^{-5}$), with 16 rescues and no observed harms. Its numeric-suffix-cluster sensitivity interval is [5.2, 15.4]. Pooling the completed-primary sensitivity and extension gives 149/320 (46.6\%) versus 186/320 (58.1\%), a gain of 11.6 percentage points. The pooled task-bootstrap interval is [8.1, 15.3], and the cluster sensitivity interval is [7.0, 16.4]. The pooled transitions are 149 tasks successful under both conditions, 37 rescues, no observed harms, and 134 tasks unsuccessful under both (Table~\ref{tab:main-results}).

The pooled gain is positive in all four categories (Table~\ref{tab:category}). Section~\ref{sec:attribution} traces these gains to a shared repair pattern.

\subsection{Intervention selectivity and failure ranking}
\label{sec:ranking}

On the completed-primary sensitivity, the complete policy selects 39 of 88 actor failures and none of 72 actor successes, giving selection precision 1.000, recall 0.443, and false-positive rate 0.000. On the extension, it selects 32 of 83 failures and none of 77 successes, giving precision 1.000, recall 0.386, and false-positive rate 0.000. Pooled recall is 0.415. These statistics measure selection of actor failures; successful repair is a separate outcome.

The policy accepts 71 interventions across 3,665 recorded action decisions (1.94\%). Of these, 37 rescue a failed trajectory and 34 leave it unsuccessful, for a repair yield of 52.1\%. All 149 actor-successful trajectories are preserved without intervention. The absence of regressions establishes preservation on this set, but does not measure the safety of modifying a successful trajectory.

\paragraph{Failure ranking.}
For failure ranking, define $D_j=\max_{t\leq\tau_j} d_{j,t}$, where $\tau_j$ is the first accepted replacement step or the final step if no replacement occurs. The score uses only the replayed actor prefix, before any repair outcome. Figure~\ref{fig:detector} compares this ranking with the complete policy's intervention selection. ROC-AUC is 0.753 on the completed-primary sensitivity and 0.768 on the conditional extension; pooled ROC-AUC is 0.759 (numeric-suffix-cluster 95\% CI [0.686, 0.824]). Average precision is 0.830 and 0.833 on the two sets, respectively, and 0.829 when pooled, compared with a pooled actor-failure prevalence of 0.534. These descriptive ranking results do not establish an incremental benefit from verification.

\begin{table}[htbp]
\centering
\caption{Trigger-matched controls on the completed-primary sensitivity. Intervals are 95\% CIs from category-stratified paired bootstrap resampling of tasks.}
\label{tab:baselines}
\small
\begin{tabular}{lrrr@{\hspace{0.8em}}rr}
\toprule
Method & Success (\%) & $\Delta$ & 95\% CI & Accepted & Rescue / harm \\
\midrule
Actor & 45.0 & -- & -- & -- & -- \\
Selective self-revision & 46.2 & +1.2 & [0.0, 3.1] & 9 & 2 / 0 \\
Selective absolute verifier & 46.9 & +1.9 & [0.0, 4.4] & 4 & 3 / 0 \\
\method (full protocol) & \best{58.1} & \best{+13.1} & \best{[8.1, 18.8]} & 39 & \best{21 / 0} \\
\bottomrule
\end{tabular}
\end{table}

\subsection{Candidate provenance and component attribution}
\label{sec:attribution}

All 71 constructed twins are deterministic: 70 use inverse-action repair and one uses a stale-argument substitution (Table~\ref{tab:provenance}). No model-generated twin is constructed on either test set. The 70 inverse repairs all remove \texttt{releaseBrakePedal}: 63 replace a single-call proposal with the fixed completion response, and seven retain other calls in the bundle. These repairs produce all 37 rescues. The gain is therefore concentrated in a vehicle-control failure pattern shared across categories; the category breakdown does not establish diversity of repaired environments.

\begin{table}[htbp]
\centering
\caption{Executed repair provenance across the completed primary set and conditional extension. Counts are reconstructed from the archived decision logs.}
\label{tab:provenance}
\small
\begin{tabular}{lrrrr}
\toprule
Repair family & Completed primary & Extension & Rescue & Harm \\
\midrule
Inverse-action (deterministic) & 38 & 32 & 37 & 0 \\
Stale argument (deterministic) & 1 & 0 & 0 & 0 \\
Model-generated alternative & 0 & 0 & 0 & 0 \\
\midrule
Total & 39 & 32 & 37 & 0 \\
\bottomrule
\end{tabular}
\end{table}

\paragraph{Fixed-candidate ablations.}
The verifier accepts all 71 structurally valid, nonidentical candidates presented to it. Removing its acceptance conditions and numerical score gate retrospectively leaves all 3,665 replacement decisions and selected actions unchanged. The evidence rules, candidates, structural checks, trajectory budgets, and stored continuations remain fixed. Removing the reverse-orientation checks also changes no decision (Table~\ref{tab:ablations}). These ablations reuse the archived candidates and continuations without resampling the model. The forward-only comparison retains the forward judgment from the original two-orientation request; it does not test a separately prompted single-order judge.

\begin{table}[htbp]
\centering
\caption{Fixed-candidate decision ablations on 320 task pairs. Identical selected actions permit reuse of the same realized continuations and outcomes; rows are not independent experiments.}
\label{tab:ablations}
\small
\begin{tabular}{lrrrr}
\toprule
Decision rule & Switches & Changed decisions & Success (\%) & Rescue / harm \\
\midrule
Full policy & 71 & -- & 58.1 & 37 / 0 \\
Without verifier or score gate & 71 & 0 / 3,665 & 58.1 & 37 / 0 \\
Stored forward orientation only & 71 & 0 / 3,665 & 58.1 & 37 / 0 \\
\bottomrule
\end{tabular}
\end{table}

\paragraph{Decision funnel and overhead.}
Of 3,665 logged decisions, 2,192 fail the evidence prefilter, 1,119 are final responses or clarifications, and 283 occur after the trajectory's replacement budget is exhausted. The remaining 71 produce nonidentical candidates; none is rejected by a structural check, verifier clause, or numerical threshold. Each uses one verifier API call containing both orientations, with mean overhead of 2,959.8 input tokens, 513.3 output tokens, and 9.22 seconds per intervened decision (median 8.81; 95th percentile 12.40 seconds). These measurements exclude local rule-processing time and downstream continuation costs (Appendix~\ref{app:audit}).

\Needspace{7\baselineskip}
\paragraph{Policy controls.}
Selective self-revision and the absolute verifier improve success by 1.2 and 1.9 percentage points, compared with 13.1 for the full policy (Table~\ref{tab:baselines}). These controls generate their own candidates and use different acceptance rules, so the comparison cannot isolate the benefit of pairwise verification. With candidates held fixed, the audit attributes the observed decisions to evidence-constrained candidate construction: the verifier adds no filtering on these trajectories.

\section{Limitations and broader impacts}
\label{sec:limitations}

\paragraph{Scope and attribution.}
The results are limited to one actor configuration and a shared vehicle-control repair pattern under BFCL's official state- and response-based checks. Task-ID separation and cluster-resampling sensitivities do not establish scenario independence or remove the extension's outcome-dependent selection. With one stored trajectory per task, we do not measure robustness to model/API resampling. A shared actor/verifier model family may also produce correlated errors. Broader candidate sets and environments are needed to establish repair coverage and verifier selectivity; the exploratory follow-up does not provide an independent test of either.

\Needspace{5\baselineskip}
\paragraph{Potential impact.}
Selective action repair may reduce erroneous state changes and silent agent failures. High-impact deployments should preserve audit trails, restrict tool privileges, and require external confirmation and independent monitoring. \method is a repair layer, not authorization for autonomous consequential action.

\section{Conclusion}

Repair at the execution boundary is asymmetric: declining a replacement preserves the actor's proposal, whereas an unnecessary replacement can create a new failure. \method addresses this risk by constraining candidate construction with trace evidence and comparing each candidate with the actor's proposal. Under exact replay, the policy improves BFCL task success without observed success-to-failure regressions. The audit attributes these gains to a shared inverse-action repair pattern; removing the verifier checks leaves every replacement decision unchanged on these trajectories.

\bibliographystyle{plainnat}
\bibliography{references}

\clearpage
\appendix

\section{Replay validation and component audit}
\label{app:audit}

\paragraph{Replay and ablation procedure.}
We check 3,382 actor actions through the first accepted replacement against the stored actor replay, verify the switch index and selected candidate on all 71 intervened trajectories, and compare the entire evaluator-visible result on all 249 nonintervened trajectories. No mismatch occurs. For each logged decision, we recompute the weighted quality from the rubric scores and compare the full acceptance rule with (i) acceptance after evidence and structural checks alone and (ii) the stored forward-orientation rule. We verify identical selected actions on all 3,665 decisions before reusing the stored continuations and official task outcomes. Restricting the anomaly-score maxima to the replayed actor prefix through the first accepted replacement leaves every ranking score unchanged. This audit uses stored decisions; it does not resample a model, rerun the environment, or estimate latency for a separately executed ablated system.

\begin{table}[htbp]
\centering
\caption{Logged decision funnel. The first four disposition rows partition all decisions. Structural and verifier rows then track the same constructed candidates.}
\label{tab:funnel}
\small
\begin{tabular}{lrrr}
\toprule
Stage or disposition & Completed primary & Extension & Pooled \\
\midrule
All recorded decisions & 1,879 & 1,786 & 3,665 \\
Evidence prefilter skips & 1,140 & 1,052 & 2,192 \\
Final-response / clarification skips & 560 & 559 & 1,119 \\
Post-switch budget skips & 140 & 143 & 283 \\
Candidates constructed & 39 & 32 & 71 \\
\midrule
Nonidentical and structurally valid & 39 & 32 & 71 \\
Pairs presented to verifier & 39 & 32 & 71 \\
Both orientations pass all clauses & 39 & 32 & 71 \\
Accepted replacements & 39 & 32 & 71 \\
\bottomrule
\end{tabular}
\end{table}

No candidate fails schema validity, permitted action-kind/call-count changes, edit-distance, grounding, or nonidentity checks. At judgment, none fails twin preference, actor-fatal labeling, twin-nonfatal labeling, confidence $0.98$, quality margin $0.50$, or the calibrated $0.50$ threshold. These zero counts describe the observed candidate set; they do not establish that any check can discriminate valid from invalid repairs.

\paragraph{Illustrative rescued trajectory.}
In \nolinkurl{multi_turn_base_62}, the user asks to start the vehicle engine to check fuel. The trace records a successful \texttt{pressBrakePedal} followed by \texttt{startEngine}, which reports a running engine. The actor then proposes \texttt{releaseBrakePedal}. The inverse-action repair removes that call and emits the fixed completion response. The official evaluator records actor failure and wrapper success: the actor's brake state is \texttt{released}, while the reference state is \texttt{pressed}; the engine is running in both. The repair preserves the benchmark's reference state. This does not establish that releasing a brake is universally unsafe or inconsistent with a natural-language request.

\clearpage
\section{Illustrative certificate-to-twin constructions}
\label{app:certificate-examples}

\begin{figure}[htbp]
  \centering
  \includegraphics[width=\linewidth]{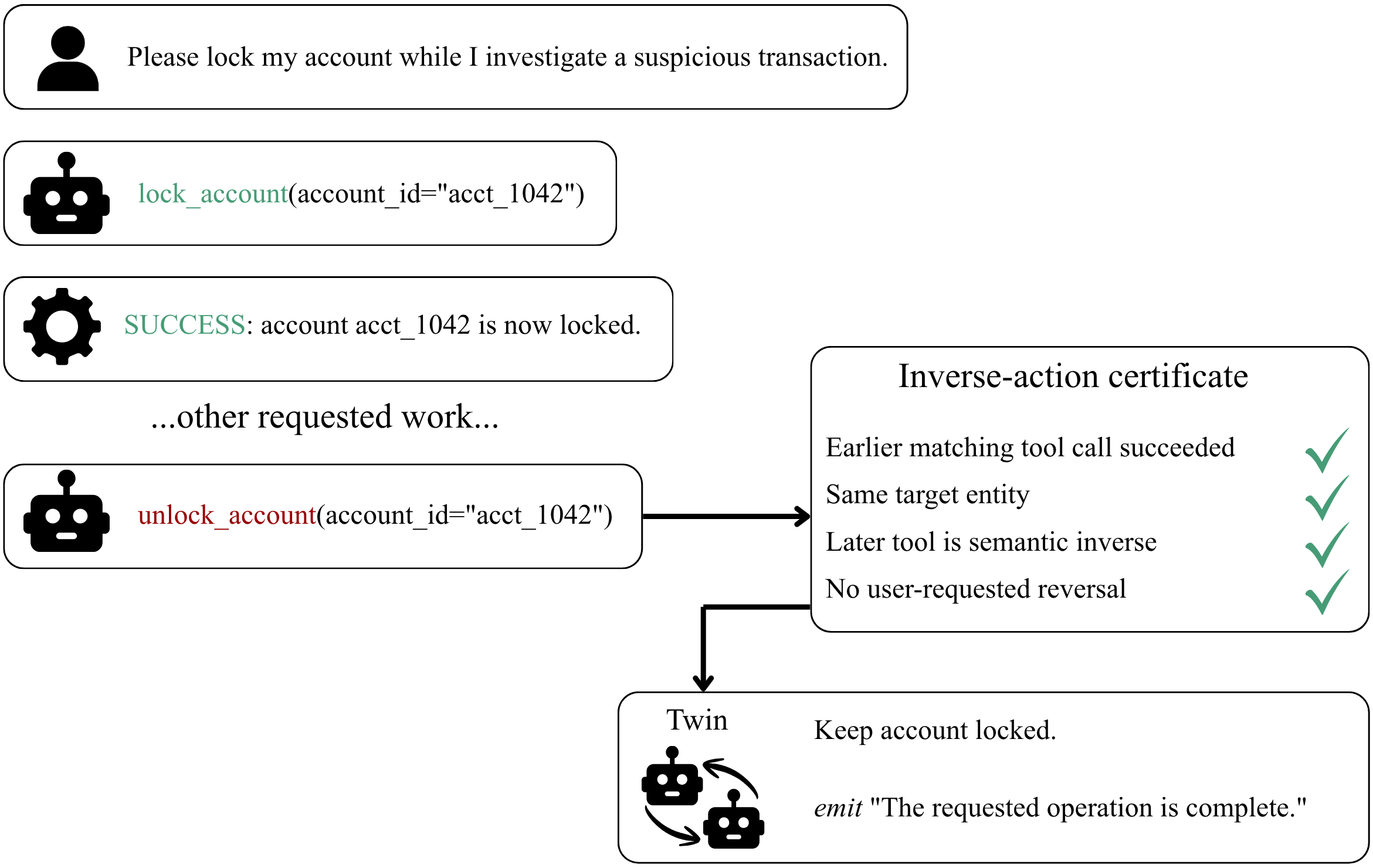}
  \caption{Schematic lexical inverse-action construction. Following a successful \texttt{lock\_account} call, the actor proposes \texttt{unlock\_account} on the same account without a user-requested reversal. The fixed-table inverse match authorizes deletion of that call. Because the trace contains a successful operation and no unresolved later failure, the deterministic twin is the fixed completion response. The phrase ``semantic inverse'' in the diagram denotes this prespecified lexical relation, not open-ended semantic inference.}
  \label{fig:inverse-action-example}
\end{figure}

\begin{figure}[htbp]
  \centering
  \includegraphics[width=\linewidth]{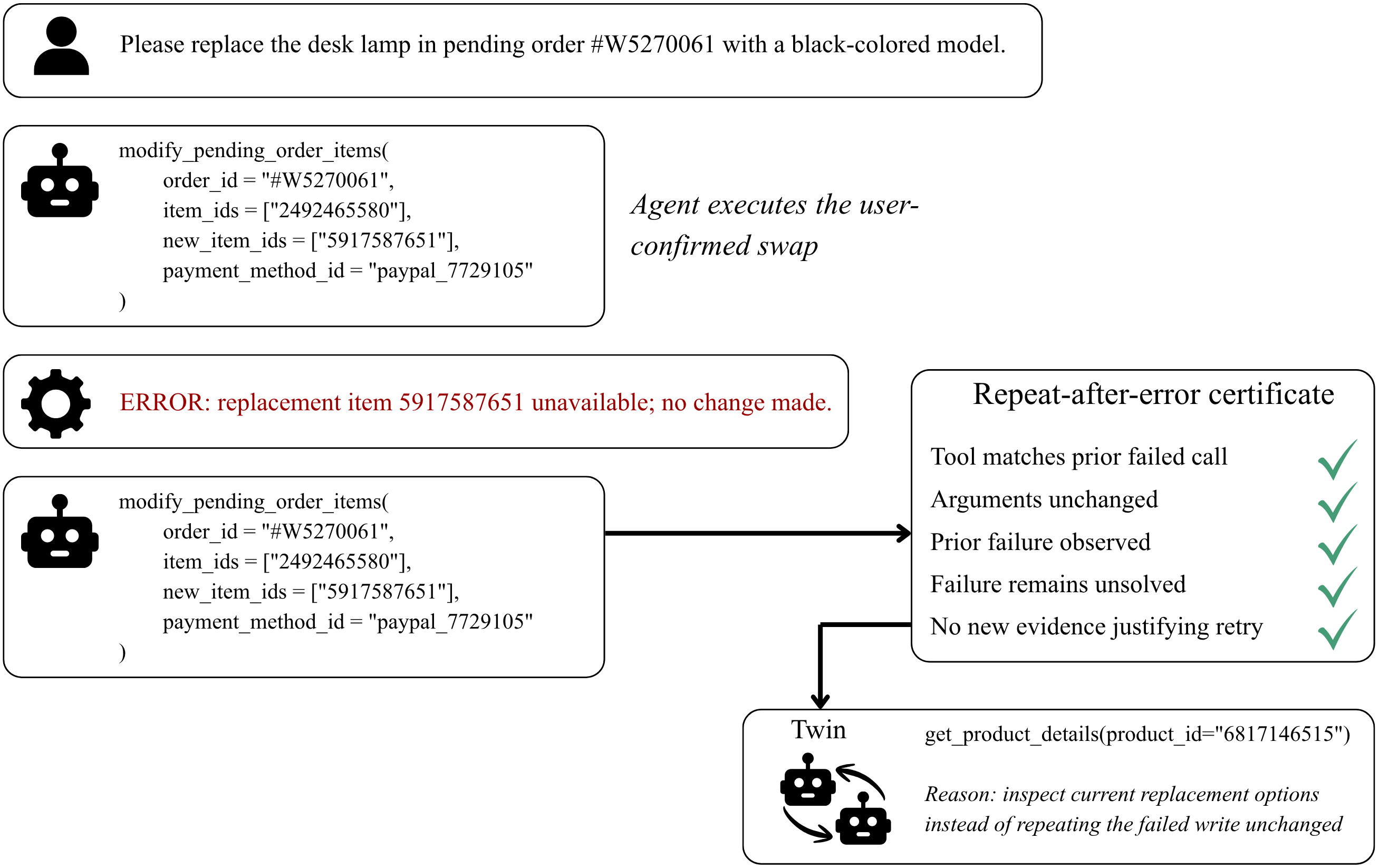}
  \caption{Schematic repeat-after-error construction adapted from a retail interaction pattern. A modification fails, and the actor repeats the same tool with identical arguments. Because no later successful current-turn tool record intervenes, the certificate authorizes structural change. The generator proposes an evidence-gathering call instead of reissuing the failed write. The queried product identifier is established in trace context omitted for space; without this grounding, structural validation would reject the candidate.}
  \label{fig:repeat-after-error-example}
\end{figure}

\begin{figure}[htbp]
  \centering
  \includegraphics[width=\linewidth]{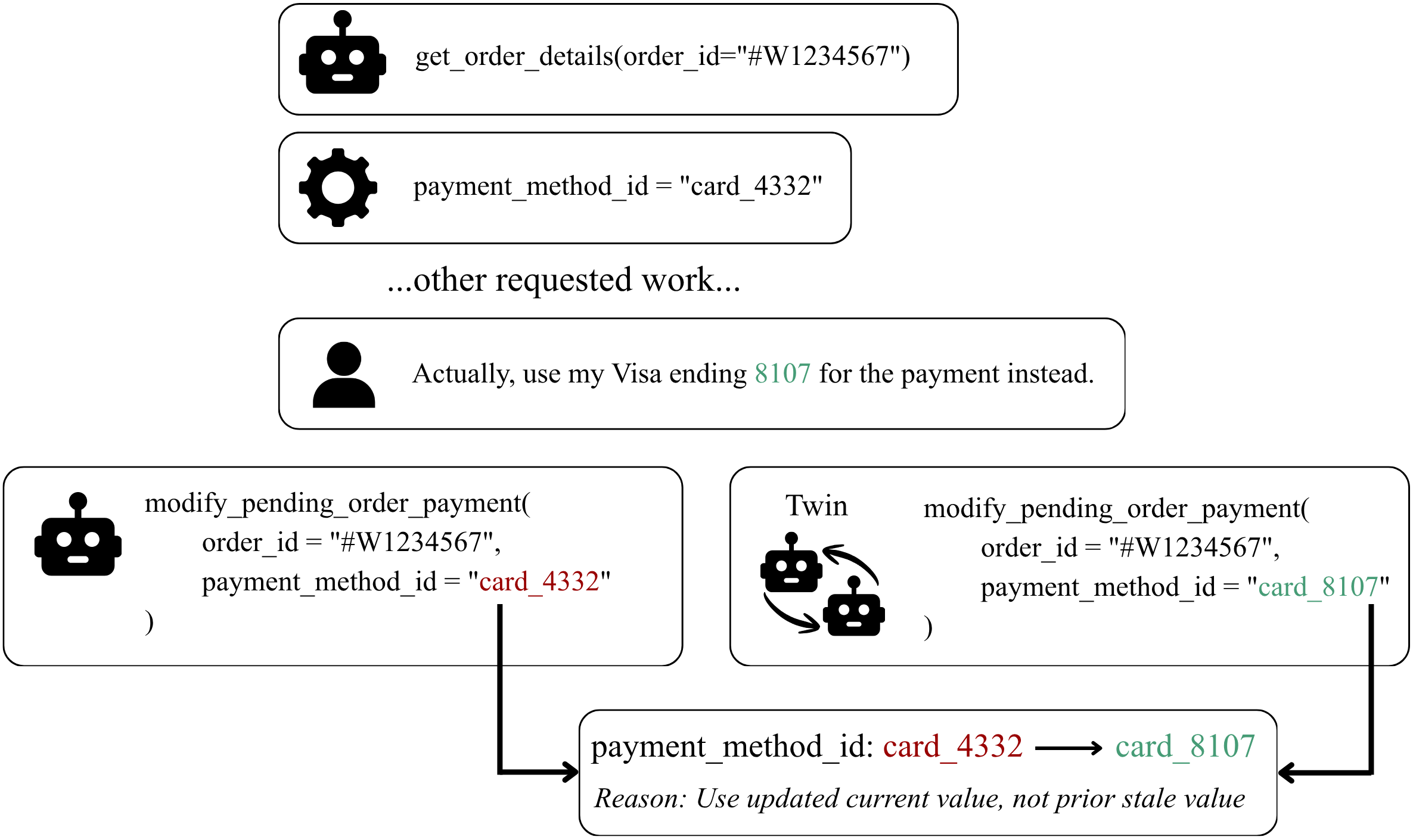}
  \caption{Schematic unique-stale-argument construction adapted from a retail interaction pattern. An earlier lookup supplies \texttt{card\_4332}, but the latest user message uniquely requests the card ending in \texttt{8107}. Context omitted from the diagram supplies the corresponding identifier, \texttt{card\_8107}. The actor carries forward the stale identifier. The deterministic twin replaces only \texttt{payment\_method\_id}, preserving the rest of the call.}
  \label{fig:stale-argument-example}
\end{figure}

Figures~\ref{fig:inverse-action-example}--\ref{fig:stale-argument-example} illustrate the three evidence conditions in Section~\ref{sec:method}. Figure~\ref{fig:inverse-action-example} is a synthetic account-control example. Figures~\ref{fig:repeat-after-error-example} and~\ref{fig:stale-argument-example} are manually constructed composites adapted from interaction patterns and tool interfaces in the Retail domain of $\tau$-bench~\citep{yao2024tau}. Identifiers and incidental trace details are simplified or altered for exposition; these are not sampled $\tau$-bench trajectories or additional evaluation results. The annotations explain each certificate, while the operational predicates in Section~\ref{sec:method} determine eligibility. Each panel shows candidate construction. Execution still requires a nonidentical candidate to pass the structural checks and counterbalanced verifier gate. All quantitative claims concern the BFCL V4 evaluation.

\clearpage
\section{Exploratory case study on broader candidate selection}
\label{app:followup}

These studies ask whether a selector can distinguish improvements from regressions when candidate quality varies. They use broader generators or revised selection rules and leave the primary algorithm and results unchanged. All task IDs had prior exposure, and revisions followed failure analysis. The results are therefore exploratory development evidence. Rates are pooled task proportions rather than the category-balanced primary statistic.

\paragraph{Matched-candidate design.}
The 361-task cohort contains the original 401 development IDs minus 40 follow-up pilot IDs. A proposer guided by user requirements and their completion status examined at most eight actor decisions per task and retained the first structurally valid repair, producing 195 candidates from 1,983 attempts. Each candidate was judged before execution, then executed regardless of the verdict, followed by one sampled actor continuation without further repairs. For each task, all policies use the same recorded baseline if they reject or the same candidate continuation if they accept. Tasks without candidates remain in the denominator. This design isolates selection on fixed candidates and continuations; it does not test robustness across continuation samples or repeated repairs.

The follow-up uses the same \texttt{gpt-5.6} alias (reported as \texttt{gpt-5.6-sol}), with high, low, and medium reasoning effort for the actor, proposer, and judges. Judges compare anonymous actions A and B in both orders using the visible prefix and tool definitions, without outcomes or future observations. We compare four selectors: \emph{conservative obligations}, \emph{indexed evidence}, \emph{relative preservation}, and \emph{local correction}.

\begin{table}[htbp]
\centering
\caption{Exploratory matched-pool comparison on all 361 tasks. Accepted counts refer to the same 195 candidates; gains and losses are paired official-checker outcomes relative to the actor. They are not semantic-safety labels.}
\label{tab:exploratory-matched-pool}
\small
\begin{tabular}{lrrr}
\toprule
Policy & Successes (\%) & Accepted & Gains / losses \\
\midrule
Actor & 174 (48.20) & --- & --- \\
Apply every candidate & 180 (49.86) & 195 & 16 / 10 \\
Conservative obligations & 181 (50.14) & 36 & 7 / 0 \\
Relative preservation & 182 (50.42) & 71 & 8 / 0 \\
Local correction & 183 (50.69) & 83 & 9 / 0 \\
\bottomrule
\end{tabular}
\end{table}

\paragraph{Selection on mixed candidates.}
Unlike the primary audit, this pool contains both improving and degrading repairs (Table~\ref{tab:exploratory-matched-pool}). The final local-correction selector retains nine of the pool's 16 checker-improving continuations and rejects all ten checker-degrading continuations, forgoing seven improvements. It improves on applying every candidate by three tasks (+0.83 percentage points) and on the actor by nine (+2.49 points). In its prespecified comparison, the conservative selector improved on applying every candidate by one task. The final selector exceeds the conservative selector by two tasks (+0.55 points; four gains and two losses). These comparisons are descriptive: both task exposure and method selection preceded evaluation.

\paragraph{Selector variants.}
The indexed-evidence selector replaced quotations with source IDs and separated prefix completion from action preservation. It yielded 178 successes with 28 acceptances: valid provenance did not ensure consistent obligation annotations. Relative preservation improved treatment of completed and still-pending requirements. The final local-correction rule prevents a negative progress annotation from vetoing an otherwise supported correction. It retains every other predicate, requires documented actor errors and a clean candidate assessment, and preserves unaffected tool calls in order and multiplicity. Replacements cannot cross unaffected calls; corrections to textual actor actions remain textual. Both candidate orders must pass.

The relative-preservation and local-correction selectors reuse the indexed selector's 390 judgments; the conservative selector uses its earlier judgments. Rules were frozen before application to the complete cached pool, and final decisions were written and hashed before score aggregation. This fixes decisions within an application but does not remove prior exposure. The final application used no additional provider calls, proposals, or continuations. Its incremental reevaluation cost was zero; an online deployment would still incur inference overhead.

\paragraph{Additional candidates and citation recovery.}
Two extensions examined opportunity coverage and citation syntax. A search extension started from a policy with 182 successes, retained all 361 tasks, and considered at most two additional baseline positions. It made 308 attempts and accepted 11 new repairs, with no incremental checker gains or losses. A subsequent schema-citation recovery step reconsidered cached candidates that failed quotation-format validation while retaining the other checks. It judged 49 candidates and accepted 19, yielding 181 successes: no gains and one loss relative to the preceding policy. Together, these extensions used 635 new requests and regenerated no actor baselines. Their interventions are not pooled into the matched comparison.

\paragraph{Separate 320-task re-evaluation.}
An earlier study revisited the original 320 primary/extension task IDs with fresh actor trajectories and broader proposals. Its task IDs are disjoint from the 361-task cohort, and its actor baseline differs from the original pooled result. On a fixed pool of 76 candidates and continuations, applying every candidate raised success from 147/320 to 180/320, with 35 gains and two losses. An audit found that the original judges had received construction metadata that could be mistaken for user constraints. Fresh neutral judgments of every pair, with those labels removed, accepted only three candidates and scored 146/320: they rejected all 35 gains, prevented one loss, and accepted the other. The metadata-bearing gate scored 163/320, which does not support the neutral gate. Because metadata removal and fresh judgment were bundled, their individual effects are not isolated. All 35 gains concerned the same brake-release inverse pattern. They support candidate repair on that pattern, without establishing broad coverage or a benefit from the tested gate.

\paragraph{Checker outcomes and semantic quality.}
The final selector's one additional rescue over relative preservation corrects 750 miles to the tool-documented 750 kilometers. The correction is supported, but the official rescue occurs through later brake-state behavior in the stored continuation; the wording change alone is not established as the cause. Conversely, the citation extension's sole loss replaces an insurance purchase with a clarification because coverage and premium are distinct quantities. A checker loss need not imply a worse semantic decision. Manual review of all 12 newly admitted final candidates also found unnecessary clarification and attribution of a booking invoice specifically to an insurance purchase. Zero observed checker losses therefore does not establish an absence of semantic defects.

Local action quality, subsequent actor behavior, and checker requirements can diverge. Shared templates, scripted follow-ups, single sampled continuations, and outcome-informed development limit these findings to a modest selection result. General semantic safety remains unestablished.

\end{document}